\documentclass[10pt, conference, compsocconf]{IEEEtran}
\usepackage[T1]{fontenc}
\usepackage[utf8]{inputenc}

\usepackage{subfigure}
\usepackage{epsfig}
\usepackage{algorithmic}
\usepackage{color}
\usepackage{latexsym}
\usepackage{amssymb}
\usepackage{verbatim}
\usepackage[nospace]{cite}

\usepackage[skip=10pt,font=footnotesize]{caption}

\usepackage{amssymb,amsmath}
\usepackage[ruled,vlined]{algorithm2e}

\newtheorem{Lemma}{Lemma}

\newtheorem{Definition}{Definition}

\newcommand{\Comment}[1]{}

\newenvironment{myalign}{\par\nobreak\small\noindent\align}{\endalign}

\def\IEEEbibitemsep{.20\baselineskip plus 0.20\baselineskip minus
0.15\baselineskip}
\renewcommand{\IEEEQED}{\IEEEQEDopen}

\begin{document}
%

\title{Outlier Detection from Network Data with Subnetwork Interpretation}

\author{\IEEEauthorblockN{Xuan-Hong Dang, Arlei Silva, Ambuj Singh}
\IEEEauthorblockA{University of California Santa Barbara\\
\{xdang,arlei,ambuj\}@cs.ucsb.edu}
\and
\IEEEauthorblockN{Ananthram Swami}
\IEEEauthorblockA{Army Research Laboratory\\
ananthram.swami.civ@mail.mil}
\and
\IEEEauthorblockN{Prithwish Basu}
\IEEEauthorblockA{Raytheon BBN Technologies\\
pbasu@bbn.com}}

\maketitle

\begin{abstract}

Detecting a small number of outliers from a set of data observations is always challenging. This problem is more difficult in the setting of multiple network samples, where computing the anomalous degree of a network sample is generally not sufficient. In fact, explaining why the network is exceptional, expressed in the form of subnetwork, is also equally important. In this paper, we develop a novel algorithm to address these two key problems. We treat each network sample as a potential outlier and identify subnetworks that mostly discriminate it from nearby regular samples. The algorithm is developed in the framework of network regression combined with the constraints on both network topology and L1-norm shrinkage to perform subnetwork discovery. Our method thus goes beyond subspace /subgraph discovery and we show that it converges to a global optimum. Evaluation on various real-world network datasets demonstrates that our algorithm not only outperforms baselines in both network and high dimensional setting, but also discovers highly relevant and interpretable local subnetworks, further enhancing our understanding of anomalous networks.

\end{abstract}

\maketitle

\vspace{-2ex}

\section{Introduction}                                    \label{sec:Intro}

Detecting and characterizing exceptional patterns is an important task in many domains ranging from fraud detection, environmental surveillance, to various health care applications~\cite{Arthur12,Leman15}. This problem is often referred to as \textit{outlier} or \textit{anomaly} detection in the literature. In contrast to other popular data mining tasks like clustering, classification or frequent patterns mining that all discover prevalent patterns, outlier identification aims at uncovering a small set of inconsistent objects (\textit{outliers}) that deviate significantly from the larger number of regular objects (\textit{inliers}) in the data.

Although identifying anomalous subjects has been widely studied in high dimensional data~\cite{Arthur12} and recently extended to the network context~\cite{Leman15}, the problem remains very challenging. One of the most challenging issues lies in the fact that the number of anomalous objects is considerably smaller than the large population of regular ones, which limits the learning capability of most data mining algorithms. Another challenge comes from the notion of ``inconsistency'' which is hard to precisely define, quantify and interpret, especially when entities are connected in a network. In the network setting, 
most existing works focus on searching individual nodes~\cite{Henderson11}, or groups of linked nodes~\cite{EberleH07} whose structures or behaviors are irregular. Though these studies have provided intuitive concepts about outlying patterns defined in the respect of network connectivity, most results are limited to the setting of a single static network. Other recent studies have extended the scope of analysis to evolving networks~\cite{Netspot13,GuptaGSH12}, but the focus is on event/change detection where the temporal dimension is a key factor for defining outliers.

In this paper, we address the problem of identifying anomalous networks from a database of multiple network samples while at the same time investigating \textit{why} a network is exceptional. An outlier is defined at the global level of an entire network sample but we use local subnetworks to explain its exceptionality. Although the outlierness of a network sample can be quantified via the outlier degree, such a single measure only bears limited explanatory information~\cite{Barbora13,Dang14} since it lacks the capability of showing in what data view, i.e. local subnetworks, an anomalous network is most exceptional. 
Moreover, although two networks may have similar outlier degrees, the local subnetworks that make them abnormal might be quite different since the anomalous networks themselves are usually not homogeneous. For example, exploring a database of gene networks for outliers can lead to the isolation of subjects suffering from cancer. However, the gene pathway (local subnetwork) that causes the disease can vary from subject to subject due to the complexity of the disease~\cite{Dong07}, or even depending on different stages of the disease. Spotting an unhealthy subject is generally not sufficient. Figuring out what abnormal gene subnetwork leads to the disease is usually more important since it helps to develop possible and effective treatments.

We develop a novel algorithm that exploits network regression models combined with network topology regularization to concurrently address the two important problems mentioned above. Specifically, we treat each network sample as a potential outlier and determine local subnetworks that help discriminate it from nearby regular network samples. Our objective function is formulated under the framework of network regression where we first upsample the outlier candidate network in order to make the binary regression problem balanced. The objective function is then regularized by the network topology and further penalized by L1-norm shrinkage to perform subnetwork discovery. It can be shown that the combined objective function has a form closely related to the dual SVM~\cite{Hsieh08,HasTibFri09}, which can be further optimized in the primal form using Newton's method. The objective function is proven to be convex, which is key to guaranteeing the convergence of the algorithm. Our algorithm, therefore, goes beyond the simple strategy of subspaces/subgraphs examination by directly learning the most discriminative subnetworks with respect to each network sample. Consequently, the outlier degree can be appropriately computed within the space spanned by these selected subnetworks and, collectively, they form a ranking of all network samples based on the outlier scores. 

In summary, we make the following contributions in this work: (i) We address a challenging problem of both identifying and explaining anomalous networks from a database of network samples. The explanations are expressed in the form of local subnetworks, which play a key role in understanding the abnormal properties behind the observed network data; (ii) We formulate the problem under the regression framework with network regularization for subnetwork discovery, and develop a novel algorithm to efficiently mine most relevant subnetworks to discriminate and explain network outliers from their nearby network inliers; (iii) We demonstrate the effectiveness of our algorithm against typical techniques developed for both dynamic network data and high dimensional data using various real world datasets.
Experimental results show that our algorithm is not only competitive in producing outlier ranking quality but further outputs highly relevant and interpretable local subnetworks, leading to better understanding of why the outlier networks are exceptional.

\section{Problem Setting}                    \label{sec:Preliminaries}

\begin{Definition} \label{Def1}
\noindent 
A \emph{network sample} is a triple $\mathcal{N}_k=(\mathcal{V}_k,E_k,\mathcal{F})$, where $\mathcal{V}_k=\{v_1,v_2,\ldots,v_{n}\}$ is a set of nodes, $E_k\subseteq \mathcal{V}_k\times \mathcal{V}_k$ is a set of undirected edges, and $\mathcal{F}$ is a function labeling each node with a real number.
\end{Definition}
\vspace{1ex}

Let $\mathcal{DB} = \{\mathcal{N}_1, \mathcal{N}_2, \ldots,\mathcal{N}_m\}$ be a network dataset that consists of $m$ network samples. We focus on a family of networks whose topologies are relatively stable across different network instances. For example, human subjects usually have similar gene networks with the same number of genes. However, the expression level of each individual gene may differ from subject to subject. Likewise, various snapshots captured from a traffic network often have the same network topology while traffic conditions on each road segment may vary from snapshot to snapshot. In mining outlying networks from a database of network samples $\mathcal{DB}$, we aim to compute an anomaly score for each network sample and at the same time, to uncover subnetworks that show the most exceptional properties of the network under examination. Collectively, an outlier ranking is generated for the entire dataset and those network samples having the highest anomaly scores will be brought up to the user for further investigation.

\section{Regression on Networks}   \label{sec:RGNN}

\Comment{
Identifying outlier subjects often relies upon notion of distance which generally quantifies the deviation of a given subject from the global mean of all subjects. Such a distance can be measured in either the original data space or the subspace spanned by the dominant eigenvectors corresponding to the largest eigenvalues. Such an approach generally implies that both inliers and outlier subjects are homogeneous
This addresses for the most general case in which the network distribution can be complex and may contain multiple groups of common network samples. 

Analyze:

- high dim, many irrelevant nodes?

- heterogeneous data, no assumption of data distribution, thus emphasize local outliers.

- how to put linear regression into the context: allow us to perform feature selection through regularization.

- aggregate network is defined as the supergraph rather than the average one as we also want to find outlier subnetwork from topology aspect as well.

Both outlier score and the quality of explanatory subnetworks are thus dependent on how we take the network data and their structure into account. Unlike conventional approaches from unstructured data that broadly rely upon the notion of distance of one subject to the global mean of all subjects~\cite{PINN}, we make no assumption about the data distribution but focus on the local neighborhood structure of each potential outlier network. 

}

As mentioned in the previous section, our objective is not only to compute the outlier degree for each network sample but also to discover a small set of subnetworks as explanations for each outlier candidate network. We explore the regression model for our problem since it allows us to formulate outlier detection as a binary prediction. In this section, we formulate the regression problem solely based on the values associated with network's nodes while the network topology will be taken into account in the next section. 

We view each network sample as a potential candidate outlier while comparing its properties against its $K$ nearby networks (based on some network distance measures, e.g. cosine distance between node values~\cite{SNL}).
Therefore, a network sample can be a \textit{local} outlier rather than a \textit{global} one~\cite{Arthur12,LOF}, as both network distribution and the outliers themselves can be heterogeneous and one should not presume any canonical form for the distribution. Let us denote $\mathcal{N}_o$ as an outlier network candidate, and 
$\mathcal{N}_k$ as one of its $K$ neighboring networks (we use the same index $k$ as in Def.\ref{Def1} for simplification, but here $k$ only ranges over the $K$ nearest neighbors of $\mathcal{N}_o$). We can capture the node-values of a network sample $\mathcal{N}_k$ by a vector $\mathbf{x}_k$ in a high dimensional space $\mathbb{R}^n$. Under the vector format, we aim to optimize the following regression function for each $\mathcal{N}_o$:  

\Comment{
Note that, in the general case, $\mathbf{x}_k$'s may be slightly different in size (e.g. missing genes in some subjects, or a few users join/leave a social network at some snapshots), but they are all still a subset of $\mathbb{R}^n$, and we can simply use null values to denote missing nodes.}

\vspace{-1ex}
\begin{myalign} \label{obj-01}
\arg\min_{\mathbf{w}}L(\mathbf{w})\!\!=\!\! (\mathbf{x}_o^T\mathbf{w}\!\!-\!\!z_o)^2 \!+\! \sum_{k=1}^K (\mathbf{x}_k^T\mathbf{w}\!\!-\!\!z_k)^2~s.t.~ |\mathbf{w}|_1\!\!\leq\!\! 1
\end{myalign}
\vspace{-2ex}

\noindent where $\mathbf{x}_o$ is the vector of local node values for network $\mathcal{N}_o$; and $z_o=-1$ while 
$z_k=1$ if $\mathcal{N}_k$ is among the $K$ neighboring networks of $\mathcal{N}_o$; $|\mathbf{w}|_1$ is the L1-norm of vector $\mathbf{w}$. The main role of $|\mathbf{w}|_1$ is to set many coefficients in $\mathbf{w}$ to zero if the corresponding nodes are less predictive. It is worth mentioning that in a conventional case, one can constrain $ |\mathbf{w}|_1 \leq c$~\cite{HasTibFri09} for a non-negative constant $c$. However, it is easy to see that $c$ is only a scalar and can be replaced by 1 by dividing both $\mathbf{w}$ and the predicted labels $z_o$, $z_k$'s by $c$. For simplicity, we thus directly use the constraint $|\mathbf{w}|_1 \leq 1$.

\Comment{
Recall that the class of networks we are addressing have similar network structures and thus we assume that $\mathbf{x}_o$ and all $\mathbf{x}_k$'s can be represented with the same data dimensionality of $n$ nodes (the null value can be used to denote the state of a missing node in a network). 
}

It is possible to see that our Eq.\eqref{obj-01} resembles the form of Lasso regression~\cite{HasTibFri09}. However, there are two challenging issues in optimizing Eq.\eqref{obj-01}. First, our regression model is highly imbalanced since we have only a single outlier candidate but a large number of neighboring inliers. In dealing with this issue, we adopt a simple approach of upsampling the outlier candidate in order to ensure that the data become balanced\cite{batista2004study}. Essentially, $(K-1)$ new samples will be generated (for the outlier class) following the normal distribution with $\mathbf{x}_o$ as the mean vector, and the covariance matrix as the one computed from the statistics of $K$ neighboring networks. By doing so, we assure that variations at each node/dimension of the outlier class are not generated randomly 
but resemble the ones from the inlier class, and thus minimize the impact on the explanation quality of the outlier.

The second, more challenging, issue in optimizing Eq.\eqref{obj-01} is that the function is not directly differentiable---it is not smooth due to the appearance of L1-norm imposed on $\mathbf{w}$. The solution is at best only suboptimal using methods like sub-gradient~\cite{Schmidt07}, in which each component of $\mathbf{w}$ is optimized individually and in sequence. Moreover, such a solution is less efficient given the large number of nodes in the networks. We thus handle the L1-norm in a more general setting~\cite{Schmidt07} by representing $\mathbf{w}$ using two non-negative variables $\mathbf{w}^+$ and $\mathbf{w}^-$,
that are respectively defined as 
$\mathbf{w}^+ = \max(0,\mathbf{w})$ and $\mathbf{w}^- = - \min(0,\mathbf{w})$. Hence, it is easy to see that $\mathbf{w} = \mathbf{w}^+ - \mathbf{w}^-$. We denote the new variable $\tilde{\mathbf{w}} = [\mathbf{w}^+; \mathbf{w}^-] \in \mathbb{R}^{2n}$. Coefficients in $\tilde{\mathbf{w}}$ are thus \textit{all} non-negative. Now, in combination with the upsampling reasoning above, Eq.\eqref{obj-01} can be reformulated in the matrix form as follows:

\vspace{-2ex}
\begin{myalign} \label{obj-02}
\arg\min_{\tilde{w}_i \geq 0}L(\tilde{\mathbf{w}}) = \left \|[X^{(o)}, -X^{(o)}]\tilde{\mathbf{w}} - \mathbf{z}\right \|_2^2~~ s.t. \sum_{i=1}^{2n} \tilde{w}_i \leq 1
\end{myalign}
\vspace{-2ex}

\noindent where $X^{(o)}$ is the matrix with the first $K$ rows as the vectors $\mathbf{x}_k$'s, and the last $K$ rows as $\mathbf{x}_o$ and its $(K-1)$ sampling vectors. Correspondingly, the first $K$ entries of vector $\mathbf{z}$ are $+1$, predicting $\mathbf{x}_k$'s as inliers, while the last $K$ entries are $-1$, predicting $\mathbf{x}_o$ and its upsampling samples as outliers.

\section{Role of Network Topology}   \label{sec:NetTopo}

Our formulation in Eq.\eqref{obj-02} gives us a regression form to predict $\mathcal{N}_o$ as an outlier candidate based on the local state values associated with network nodes. It, however, has not taken into account the network information, and may lack essential information in learning the most relevant subnetworks that make $\mathcal{N}_o$ exceptional. 
Therefore, we add the network structure information as a constraint in learning $\mathbf{w}$'s coefficients. Intuitively, if two nodes are connected in the network, their behaviors will mutually impact each other and, consequently, their coefficients reflected in $\mathbf{w}$'s entries should be similar. For example, if a congestion happened at a road segment (node), it is likely that the nearby road segments will also be impacted, causing low speed over a region of the network. Towards modeling this network influence, we first define a graph that generalizes the network topology of both $\mathcal{N}_o$ and its $K$ neighboring networks as follows:

\vspace{1ex}
\begin{Definition} \label{Def2}
\noindent Let $\mathcal{DB}_o = \{\mathcal{N}_o,\mathcal{N}_p, \ldots, \mathcal{N}_q\}$ be the set of networks that involves the outlier candidate network $\mathcal{N}_o$ and its $K$ neighboring networks $\{\mathcal{N}_p, \ldots, \mathcal{N}_q\}$. We define $G^{(o)}=(\mathcal{V},E,A)$\footnote{The superscript $^{(o)}$ is used for $G^{(o)}$ only but it should be understood that it also applies to $\mathcal{V},E$ and $A$ since we define $G^{(o)}$ for each outlier candidate network $\mathcal{N}_o$.} as a graph summarizing the network topology of $\mathcal{DB}_o$, where $\mathcal{V}$ is the union of $\mathcal{V}_o$ and $\mathcal{V}_p,\ldots,{V}_q$;  
$E\subseteq\!\!\mathcal{V}\!\times\!\mathcal{V}$ and $E_k\!\subseteq\!E$ $\forall\! \mathcal{N}_k \in \mathcal{DB}_o$. Each edge $E(i,j)\!\in\!E$ is associated with a positive weight $A(i,j)$ defined as the popularity of the corresponding edge in either $\mathcal{N}_o$ or in its neighboring networks $\mathcal{N}_k$'s, i.e., $A(i,j)=\max(E_o(i,j),(1/K)\times\sum_k E_k(i,j))$
with $E_k(i,j)=1$ if $v_i$ connects $v_j$ in a network $\mathcal{N}_k \in \mathcal{DB}_o$.
\end{Definition}

\vspace{1ex}

We will regularize $\mathbf{w}$ using $G^{(o)}$'s topology in order to favor subnetworks that are frequently seen in $\mathcal{N}_o$ and/or in its $K$ neighboring networks, and not favor subnetworks that appear occasionally in $\mathcal{N}_k$ 's and that absent in $\mathcal{N}_o$ (i.e., $(1/K)\times\sum_k E_k(i,j))$ is small while $E_o(i,j)=0$). Values for entries in matrix $A$ are thus constrained between 0 and 1. 
Moreover, since all $\mathcal{N}_j \in \mathcal{DB}_o$ are undirected networks, $A\!\in\!\mathbb{R}^{n\!\times\!n}$ is a symmetric matrix, with $n$ as the total number of nodes in $\mathcal{V}$. 

In searching for the subnetworks that explain the abnormal properties of network sample $\mathcal{N}_o$, we impose a smoothness constraint on $\mathbf{w}$'s coefficients with respect to the graph topology captured by $G^{(o)}$. In combination with the $L1$-norm imposed on $\mathbf{w}$ (ref. Eq\eqref{obj-01}), they will together perform group/subgraph selection that predicts $\mathcal{N}_o$ as an outlier network.

Essentially, let us define the degree of a vertex $v_i$ in the graph $G^{(o)}$ as $deg(i) = \sum_{v_i\thicksim v_j} A(i,j)$, i.e. sum over all unordered pairs $\{v_i,v_j\}$ for which $v_i$ and $v_j$ are linked in  $G^{(o)}$. We assume that $G^{(o)}$ is connected (if not, each of its disconnected component will be considered separately) and thus the degree of every node is non-zero. Accordingly, the matrix $L^{(o)}$ is defined as follows:

\begin{myalign}\label{L-mat}
 L^{(o)}_{ij}=L^{(o)}_{ji}  \begin{cases}
   1-A(i,j)/deg(i) & \text{if  $v_i = v_j$}  \\
   - A(i,j)/\sqrt{deg(i) deg(j)} & \text{if $v_i$ connects $v_j$} \\
   0  & \text{otherwise }
  \end{cases}
\end{myalign}

\Comment{
\begin{myalign} \label{L-mat}
 L^{(o)}_{rs}=L^{(o)}_{sr} =
  \begin{cases}
   \sum_j A(i,j) & \text{if  $v_i \equiv v_j$} \\
   - A(i,j) & \text{if $v_i$, $v_j$ are connected in $G^{(o)}$} \\
   0  & \text{otherwise }
  \end{cases}
\end{myalign}
}

It is not hard to show that $L^{(o)}$ is positive semidefinite and it is the normalized Laplacian matrix of $G^{(o)}$. Thus, the network topology can be taken as the regularization constraint imposed on the $\mathbf{w}$ via minimizing the following quadratic form:

\vspace{-2ex}
\begin{myalign} \label{L-deploy}
\mathbf{w}^T L^{(o)} \mathbf{w} &= \sum_{v_i} \sum_{v_j}\left( \frac{w_i}{\sqrt{deg(i)}} - \frac{w_j}{\sqrt{deg(j)}}\right)^2A(i,j) \geq 0
\end{myalign}
\vspace{-2ex}

It can be seen that if $v_i$ and $v_j$ are connected in $G^{(o)}$ with a large value $A(i,j)$, the function will incur a large penalty wherever $\frac{w_i}{\sqrt{deg(i)}}$ and $\frac{w_j}{\sqrt{deg(j)}}$ are different from each other. Thus, these coefficients should be similar/smooth in order to minimize this penalty. For example, if node $v_i$ is highly explanatory for the abnormal property of $\mathcal{N}_o$, then there is a high possibility that $v_j$ is also related to the abnormality of $\mathcal{N}_o$ if both nodes are strongly connected (i.e., $w_i\approx w_j \neq 0$). Likewise, if $v_i$ is less explanatory for $\mathcal{N}_o$, its non-selection ($w_i=0$) will make $v_j$ also less likely to be selected. However, in order to appropriately incorporate this network-constrained penalty into our objective function formulated in Eq.\eqref{obj-02}, we need the following lemma.

\begin{Lemma} \label{Lemm-L}
Given the definition $\tilde{\mathbf{w}} = [\mathbf{w}^+; \mathbf{w}^-]$, the following equation is satisfied:

\begin{myalign} \label{QuadForm}
\mathbf{w}^T L^{(o)} \mathbf{w} = \tilde{\mathbf{w}}^T \begin{bmatrix} L^{(o)} & -L^{(o)} \\ -L^{(o)} & L^{(o)} \end{bmatrix} \tilde{\mathbf{w}}
\end{myalign}

\end{Lemma}

Proof: The proof of this lemma is straightforward with the expansion over the quadratic forms in both sides of Eq.\eqref{QuadForm}. 
{\hspace{1pt} \hfill $\square$ }

Lemma~\ref{Lemm-L} ensures that the network constraint penalty can also be represented using the transformed variable $\tilde{\mathbf{w}}$. Following this, we recast our objective function in Eq.\eqref{obj-02}:

\vspace{-2ex}
\begin{myalign} \label{obj-03}
\arg\min_{\tilde{w}_i \geq 0}\mathcal{L}(\tilde{\mathbf{w}}) &= \left \|[X^{(o)}, -X^{(o)}]\tilde{\mathbf{w}} - \mathbf{z}\right \|_2^2 \\[-0.5em]
&+ \lambda_1 \tilde{\mathbf{w}}^T \begin{bmatrix} L^{(o)} & -L^{(o)} \\ -L^{(o)} & L^{(o)} \end{bmatrix} \tilde{\mathbf{w}}~~s.t. \sum_{i=1}^{2n} \tilde{w}_i \leq 1\nonumber 
\end{myalign}

Notice that if the inequality constraint in Eq.\eqref{obj-03} is not equal to 1, i.e., $|\tilde{\mathbf{w}}|_1 < 1$, then the upper bound is inactive and in this case, coefficients in $\tilde{\mathbf{w}}$ will be widely non-zero. In other words, the majority of nodes in the graph will be selected. This solution is obviously undesirable. Therefore, in order to ensure that only subnetworks with the most explanatory information are used for $\mathcal{N}_o$, this constraint should always be tight~\cite{Boyd04,Zhou15}. This means that we can safely use the equality constraint $\sum_{i=1}^{2n} \tilde{w}_i = 1$, or with $\mathbf{1}$ as the vector of all 1, we have $\mathbf{1}^T\tilde{\mathbf{w}} = 1$. Upon this setting, the first term in Eq.\eqref{obj-03} can be rewritten as:

\begin{myalign} \label{term1}
&\left\|[X^{(o)},\!-\!X^{(o)}]\tilde{\mathbf{w}} - \mathbf{z}\right\|_2^2\!=\! \left\|[X^{(o)}, \!-\!X^{(o)}]\tilde{\mathbf{w}}\!-\! \mathbf{z}\mathbf{1}^T\tilde{\mathbf{w}}\right \|_2^2  \\
&= \left\|[X^{(o)} - \mathbf{z}\mathbf{1}^T, -(X^{(o)} + \mathbf{z}\mathbf{1}^T)]\tilde{\mathbf{w}}\right \|_2^2 = \left\|[X_1, -X_2]\tilde{\mathbf{w}}\right \|_2^2 \nonumber 
\end{myalign}

\noindent in which we use $X_1$ and $X_2$ to respectively denote $(X^{(o)} - \mathbf{z}\mathbf{1}^T)$ and $(X^{(o)} + \mathbf{z}\mathbf{1}^T)$. Consequently, we can combine two terms in Eq.\eqref{obj-03} into a single quadratic form by using the following lemma.

\begin{Lemma} \label{Lemm2}
Let $L^{(o)}$ be decomposed into $L^{(o)} = S^TS$ and $\tilde{X} = \left( \begin{bmatrix} X_1 \\ \sqrt{\lambda_1}S \end{bmatrix} \begin{bmatrix} -X_2 \\ -\sqrt{\lambda_1}S \end{bmatrix}\right)$. Then:
\vspace{-1ex}
\begin{myalign} \label{combinedTerm}
\!\!\!\!\left\|[X_1, -X_2]\tilde{\mathbf{w}}\right \|_2^2 + \lambda_1 \tilde{\mathbf{w}}^T\!\! \begin{bmatrix} L^{(o)} & -L^{(o)} \\ -L^{(o)} & L^{(o)} \end{bmatrix}\!\! \tilde{\mathbf{w}}
=\tilde{\mathbf{w}}^T \tilde{X}^T \tilde{X} \tilde{\mathbf{w}}
\end{myalign}
\end{Lemma}
\vspace{-1ex}

Proof: On one hand, the expansion of the first term gives us:

\begin{myalign} \label{1stTerm}
\left\|[X_1, -X_2]\tilde{\mathbf{w}}\right \|_2^2 = \tilde{\mathbf{w}}^T \begin{bmatrix} X_1^TX_1 & -X_1^TX_2\\ -X_2^TX_1 & X_2^TX_2 \end{bmatrix} \tilde{\mathbf{w}} 
\end{myalign}

On the other hand, as $L^{(o)}$ is a normalized Laplacian matrix, it can be eigen-decomposed into $L^{(o)}= U\Sigma U^T = S^TS$ where $S= \Sigma^{1/2} U^T$ where $U$ and $\Sigma$ are respectively the matrices of eigenvectors and non-negative eigenvalues of $L^{(o)}$. Therefore:

\begin{myalign} \label{combinedTerm2}
 &\tilde{\mathbf{w}}^T\begin{bmatrix} X_1^TX_1 & -X_1^TX_2\\ -X_2^TX_1 & X_2^TX_2 \end{bmatrix}\tilde{\mathbf{w}} + \lambda_1\tilde{\mathbf{w}}^T \begin{bmatrix} L^{(o)} & -L^{(o)} \\ -L^{(o)} & L^{(o)} \end{bmatrix}\tilde{\mathbf{w}} \nonumber\\
=&\tilde{\mathbf{w}}^T\begin{bmatrix} X_1^TX_1 +\lambda_1 S^TS & -X_1^TX_2 -\lambda_1 S^TS\\ -X_2^TX_1 -\lambda_1 S^TS & X_2^TX_2 +\lambda_1 S^TS \end{bmatrix}\tilde{\mathbf{w}} \nonumber \\
=& \tilde{\mathbf{w}}^T\begin{bmatrix} \begin{bmatrix} X_1 \\ \sqrt{\lambda_1} S \end{bmatrix}^T \\ \begin{bmatrix} -X_2 \\ -\sqrt{\lambda_1} S \end{bmatrix}^T \end{bmatrix} \times \begin{bmatrix} \begin{bmatrix} X_1 \\ \sqrt{\lambda_1} S \end{bmatrix} & \begin{bmatrix} -X_2 \\ -\sqrt{\lambda_1} S \end{bmatrix} \end{bmatrix}\tilde{\mathbf{w}}\nonumber \\
=&\tilde{\mathbf{w}}^T \tilde{X}^T \tilde{X} \tilde{\mathbf{w}}
\end{myalign}

\noindent From the 1st row to the 2nd row, we have used the fact that both $X_1$ and $X_2$ have the same size of $2K\times n$ while $L^{(o)}$ has the size of $n\times n$. So, the pairwise addition between the two matrices in the 2nd row is clearly matched. 
{\hspace{1pt} \hfill $\square$ }

Given Lemma~\ref{Lemm2} in combination with the previous results, we can rewrite Eq.\eqref{obj-03} as follows:

\begin{myalign} \label{obj-04}
\arg\min_{\tilde{w}_i \geq 0}\mathcal{L}(\tilde{\mathbf{w}}) = \tilde{\mathbf{w}}^T \tilde{X}^T \tilde{X} \tilde{\mathbf{w}} + \lambda_2 \tilde{\mathbf{w}}^T \tilde{\mathbf{w}} ~s.t. \sum_{i=1}^{2n} \tilde{w}_i = 1
\end{myalign}

\noindent where, like the classical ridge regression~\cite{HasTibFri09}, we add a small amount of $L2$-norm regularization in order to improve the stability of solutions when $n \gg m$. 

\section{Optimization}          \label{Sec:Opti}

In solving the objective function in Eq.\eqref{obj-04}, it is possible to note that it is closely related to the dual form of the SVM with the squared loss function~\cite{Hsieh08,SVM-review}:

\Comment{
\begin{myalign} \label{obj-dSVM0}
\arg\min_{\tilde{w}_i \geq 0} f(\tilde{\mathbf{w}}) = \sum_i\sum_j \tilde{w}_i \tilde{w}_j y_i y_j \mathbf{x}_i^T \mathbf{x}_j + \frac{1}{2C}\sum_{i=1} \tilde{w}_i^2 - \sum_i \tilde{w}_i
\end{myalign}

\noindent or in the shorter matrix form: 
}

\begin{myalign} \label{obj-dSVM}
\arg\min_{\tilde{w}_i \geq 0} f(\tilde{\mathbf{w}}) = \tilde{\mathbf{w}}^T \tilde{X}^T\tilde{X} \tilde{\mathbf{w}} + \frac{1}{2C}\sum_{i=1} \tilde{w}_i^2 - \mathbf{1}^T\tilde{\mathbf{w}}
\end{myalign}
\vspace{-2ex}

\noindent for any general dataset  $\{\mathbf{\tilde{x}}_i\}_{i=1}^{|\mathcal{DS}|}$ of $|\mathcal{DS}|$ samples, where $\tilde{X} = [\mathbf{\tilde{x}}_1,\ldots, \mathbf{\tilde{x}}_{|\mathcal{DS}|}]\times\text{diag}(\mathbf{y})$, in which $\text{diag}(\mathbf{y})$ is the diagonal matrix whose entries are class labels (i.e., $y_i \in \{-1,1\}$) for the corresponding samples $\mathbf{\tilde{x}}_i$'s, and $C$ is the margin parameter. \footnote{Note that we use the same notation $\tilde{\mathbf{w}}$ in both Eq.\eqref{obj-dSVM} and \eqref{obj-04} for easy explanation. However, $\tilde{\mathbf{w}}$ in Eq.\eqref{obj-dSVM} should be understood as the Lagrange multipliers (often denoted by $\mathbf{\alpha}$ in \cite{Hsieh08,SVM-review}). Likewise, $\tilde{X}$'s in Eq.\eqref{obj-dSVM} and \eqref{obj-04} are not necessarily the same.} 

It is easy to see that our $\tilde{X}$ in Lemma~\ref{Lemm2} can also be represented in this format. Specifically, $\tilde{X} = \left( \begin{bmatrix} X_1 \\ \sqrt{\lambda_1}S \end{bmatrix} \begin{bmatrix} X_2 \\ \sqrt{\lambda_1}S \end{bmatrix}\right)\times \text{diag}(\mathbf{y})$, where $\mathbf{y}=(1,\ldots,1,-1,\ldots,-1)^T$ is the vector in which the first $n$ entries are 1's and the last $n$ entries are -1's. Our $\lambda_2$ in Eq.\eqref{obj-04} has a similar role as $1/(2C)$ in Eq.\eqref{obj-dSVM}. The only difference between the two objective functions is that our optimization (Eq.\eqref{obj-04}) further requires the constraint $\sum_{i=1}^{2n} \tilde{w}_i=\mathbf{1}^T\tilde{\mathbf{w}}=1$. However, it also can be seen that if such a constraint is applied to Eq.\eqref{obj-dSVM}, then its last term becomes a constant. Indeed, this constraint simply rescales our optimal solution for $\tilde{\mathbf{w}}$ to be of unit L1-length. The sparseness property of $\tilde{\mathbf{w}}$ is obviously unchanged by such a normalization step. Similar to the dual-form SVM, we can solve Eq.\eqref{obj-04} using several available techniques like coordinate descent\cite{Hsieh08}, internal point~\cite{Scholkopf01} or active set method~\cite{ActiveSetSVM}. However, the computation often involves dealing with $2n$ inequality constraints directly. Therefore, a more practical approach is to consider such a quadratic programming problem in the primal form of an unconstrained problem~\cite{SVM-review,Keerthi06} as follows:

\Comment{
There are a number of methods to solve Eq.\eqref{obj-dSVM} (thus, also Eq.\eqref{obj-04}) in the dual form, including the coordinate descent\cite{Hsieh08}, internal point or active set method~\cite{}. However the computation often involves the consideration of $2m$ inequality constraints. 
}

\vspace{-1ex}
\begin{myalign} \label{obj-PF}
\tilde{\mathcal{L}}(\tilde{\mathbf{w}})\!\!=\!\!\sum_{i=1}^{2n} \sum_{j=1}^{2n} \tilde{w}_i\tilde{w}_j \tilde{\mathbf{x}}_i^T\tilde{\mathbf{x}}_j\!\!+\!\!\lambda_2 \sum_{i=1}^{2n} \max(0,1\!\!-\!\!y_i\sum_j \tilde{w}_j\tilde{\mathbf{x}}_i^T\tilde{\mathbf{x}}_j)^2
\end{myalign}
\vspace{-1ex}

\noindent where, with the introduction of vector $\mathbf{y}$ above, we have redefined $\tilde{X} \longleftarrow \left( \begin{bmatrix} X_1 \\ \sqrt{\lambda_1}S \end{bmatrix} \begin{bmatrix} X_2 \\ \sqrt{\lambda_1}S \end{bmatrix}\right)$ with $\tilde{\mathbf{x}}_i$'s as its column vectors, and $\tilde{\mathbf{w}} \longleftarrow  \text{diag}(\mathbf{y}) \times\tilde{\mathbf{w}}$. 

In this representation, one can view the first quantity in Eq.\eqref{obj-PF} as the regularization term while the second one as the loss function. Since there is a flat part in this loss function (i.e., the 2nd term in Eq.\eqref{obj-PF} is 0 if $1\!\!<\!\!y_i\sum_j \tilde{w}_j\tilde{\mathbf{x}}_i^T\tilde{\mathbf{x}}_j$), $\mathbf{\tilde{w}}$ is usually sparse. Moreover, the function is continuously differentiable, which is a great advantage. Hence, in optimizing Eq.\eqref{obj-PF}, we resort to Newton's method. Note that the function is doubly differentiable. In particular, let us denote $Q = \tilde{X}^T\tilde{X}$ and vector $Q_i$ as the $i$-th column of matrix $Q$. The gradient of $\tilde{L}$ can be written as follows:

\vspace{-1ex}
\begin{myalign} \label{gradient}
g = \frac{\partial{\tilde{L}}}{\partial{\tilde{\mathbf{w}}}} = 2Q\tilde{\mathbf{w}} - 2\lambda_2\sum_{i} Q_iy_i (1-y_iQ_i^T\tilde{\mathbf{w}})
\end{myalign}
\vspace{-1ex}

\noindent in which the summation in the second term is applied to $\tilde{\mathbf{x}}_i$'s for which $
y_i\sum_j \tilde{w}_j\tilde{\mathbf{x}}_i^T\tilde{\mathbf{x}}_j < 1$. The Hessian is therefore:

\begin{myalign} \label{Hessian}
H = \frac{\partial{\tilde{L}}}{\partial{\tilde{\mathbf{w}}}\partial{\tilde{\mathbf{w}}}^T} = 2Q + 2\lambda_2\sum_{i} y_i^2 Q_iQ_i^T
\end{myalign}
\vspace{-2ex}

At each iteration of Newton's method, we update $\tilde{\mathbf{w}}$ to $\tilde{\mathbf{w}} -\eta H^{-1}g$ where $\eta$ is the learning rate found through the line search technique\cite{Boyd04}. Given the convergence of $\tilde{\mathbf{w}}$ (thus also $\mathbf{w}$), the final subnetworks that are used as the explanations for the exceptionality of $\mathcal{N}_o$ can be identified via the non-zero entries of $\mathbf{w}$. For the outlier score of $\mathcal{N}_o$, denoted by $OS(\mathcal{N}_o)$, we follow a similar approach as~\cite{LOF} but computing it only in the subspace spanned by the explanatory subnetworks. The higher the value of $OS(\mathcal{N}_o)$, the more $\mathcal{N}_o$ deviates from its neighboring networks. 

\section{Analysis and Discussion}          \label{Sec:Discussion}

\noindent\textit{Algorithm Complexity:} We name our algorithm ODeSM that stands for \underline{O}utlier \underline{De}tection with \underline{S}ubgraph \underline{M}ining. Its complexity is briefly analyzed as follows. Searching for neighboring networks and upsampling takes $O(nm^2)$ given $m$ as the number of network samples and $n$ as the number of nodes. 
The computation of $S$ and inversion of $H$ both depend on the number of non-zero entries in $\tilde{\mathbf{w}}$ that is significantly reduced after each iteration. Let $d$ denote that number, then computing $S$ takes $O(d^2\log d)$ due to the eigen-decomposition, while the inversion of $H$ takes similar time. The checking step $(1-y_i\sum_j \tilde{w}_j\tilde{\mathbf{x}}_i^T\tilde{\mathbf{x}}_j)^2>0$ in Eq.\eqref{obj-PF}'s 2nd term takes $O(nd)$.
Due to its reliance on the Newton's method, ODeSM requires only a few iterations (usually $\leq 10$) to reach its converged solution. As this whole process is applied to each network sample, the overall complexity is therefore $O(nm^2 + m\times(d^2\log d + nd))$. 

\vspace{2ex}

\noindent\textit{Convergence:} It is straightforward to show that our Hessian matrix derived in Eq.\eqref{Hessian} is positive semi-definite. For any given non-negative vector $\mathbf{\alpha}$, we have $\mathbf{\alpha}^TH\mathbf{\alpha} \geq 0$. This is given by the fact that $Q = \tilde{X}^T\tilde{X}$, as the first term in Eq.\eqref{Hessian}, is a symmetric matrix. So its quadratic form $\mathbf{\alpha}^TQ\mathbf{\alpha}$ is always non-negative. Likewise, for the second term, each of its component's quadratic form $\mathbf{\alpha}^TQ_i Qi^T\mathbf{\alpha} \geq 0$, while $\lambda_2$ is a non-negative parameter and $y_i^2$ can be omitted as it always equals 1 by definition. These characteristics are of key importance since they collectively ensure the convexity of the objective function in Eq.\eqref{obj-PF}, making our optimization procedure always converge to the global optimal solution. Also, note that our solution lies in the general family of quadratic programming solutions, often used in both dual and primal SVM. However, unlike SVM that works in the original data space, our algorithm works in the \textit{feature} space. Nodes (or features) in the final subnetworks thus can be loosely interpreted as the (support) vectors falling inside the discriminative margin. Hence, one can control the subnetwork sizes through adjusting $\lambda_2$.

\vspace{2ex}

\noindent\textit{Parameter setting:} Other than $\lambda_2$, our algorithm requires two parameters to be set: $K$ determining neighboring networks, and $\lambda_1$ measuring the impact of network constraint. Without any prior knowledge regarding the network distribution, it is hard to choose the right values for both parameters since outlier detection is an unsupervised learning problem. We therefore employ the best-effort-approach that follows the strategy developed in~\cite{LOF,Arthur12}. The essential idea is to try on a parameter range, rather than a single value, and use an object-wise maximum ensemble to combine the final outlier score. We set $K=\{10\ldots 30\}$, similar to the range chosen in~\cite{LOF}, and $\lambda_1=\{0.1\ldots 10\}$. Regarding $K$, it is also noticed that, among neighbors of an outlier candidate, there may exist other outliers with the likelihood that they possess similar anomalous properties. In dealing with this case, one can either exclude closest neighboring samples (with assumption that outliers are closest neighbors), 
or increase the $K$ value. We have empirically tested both approaches and the results are quite similar. Indeed, since the number of outliers within a database is usually small, the probability of having one within the $K$ neighbors of a sample is usually low. The quality of the outlier detection and explanation is thus not much compromised and still determined by the majority of regular neighbors.

\section{Experiments}                                    \label{sec:Experiment}

\Comment{
3 datasets to be analyze:

	2. LA traffic data: outliers if avg speeds > a certain threshold (plot avg speed for all networks and see any broken lines to distinguish btw low/high speed)

	3. Image denoising data: having both labels and groundtruth subnetworks. For each image, position of sunglass might be different so gtruth might be varied. Note this in computing precision/recall. 
	
	4. Question about selected features? Can test on CMU where at most 70 are relevant, but for traffic data, the higher the better (reason is in the design the network, based on total average speed)

ECOutlier: java code but not for attributed graph

- Netspot: compute the anomalous value for each individual edge and form the connected subnetworks. The edge's anomalous degree is calculated based on the p-value as the fraction of network samples in which an equal and or higher weight are observed on the same edge. 
}

\subsection{Methodology}                  \label{sec:Metho}

\Comment{
We relax the temporal constraint in Netspot and set its score threshold $\Tau=...$ and the number of failures $h=...$
}

We compare the performance of our algorithm ODesM against techniques in both network studies and high dimensional studies. Specifically, it is compared against the following techniques: (1) Netspot~\cite{Netspot13} without temporal constraint so allowing it to uncover network regions from each individual network; (2) HiCS~\cite{HiCS} that seeks outliers through contrast subspaces for high dimensional data; (3) ABOD~\cite{ABOD08} which discovers outliers via variance of angles between vector triples; (4) ODesMw/o, a variant of our method that does not exploit network regularization. The parameter setting for ODesM and ODesMw/o follows the discussion in Section~\ref{Sec:Discussion}, while for Netspot, we set the number of failures $h=10$ as suggested in~\cite{Netspot13}. For HiCS, we choose all settings as suggested by the authors~\cite{HiCS} and adopt LOF as its core algorithm. ABOD is a parameter-free technique, so we use its exact version with a polynomial kernel of degree 2.

In evaluating algorithm performance, we use the well established Receiver Operating Characteristic (ROC) curve computed based on the outlier ranking returned by an algorithm against the ground truth labels of normal and outlying networks. A ROC curve provides a visualization over the relationship between the true positive rate ($y$-axis) and the false positive rate ($x$-axis). This curve
can be numerically comparable via a single value, when desired, known as the area-under-curve (AUC).

\Comment{Note: Netspot is a local search approach and thus non-deterministic. Though it has attempted to alleviate this by considering a larger set of neighboring solutions at each step./very large scale neighborhood search. HiCS: m as number of statistic tests (iteration in MC alg) = 50; } 

\Comment{
In setting the parameters for each of the technique, we vary the score threshold $\Tau$ in Netspot and $\lambda_2$ in our technique so that the subnetwork size occupies only 5\% of total network nodes. For Netspot, we further set the number of failures $h=10$ as suggested in the paper~\cite{Netspot13}. As discussed in Section~\ref{Sec:Discussion}, choosing optimal parameters for unsupervised methods is always not easy. Therefore, for every examined algorithms where applicable, we adopt the best-effort-approach to choose a realistic set of parameters. Specifically, we follow exactly the simple maximum ensemble approach as discussed for LOF in the original publication~\cite{LOF}, and later analyzed in depth in an early, model-center ensemble approach recently [3]. The range of kNN is set $K= 10\ldots30 $ while the network impact $\lambda_1= \{0.1, 0.2, 0.5,1,2.5\}$. For ABOD, we use the exact version with polynomial kernel of degree 2, while for other settings in HiCS, we use the same as suggested by the paper~\cite{HiCS} and adopt LOF as its core outlier ranking technique, since HiCS itself is already an ensemble technique. 
}

\subsection{CMUFace graph data}                  \label{sec:CMU}

\begin{figure}[t]
\hspace{-0.5cm}
\centering
\includegraphics[width=1.05\linewidth]{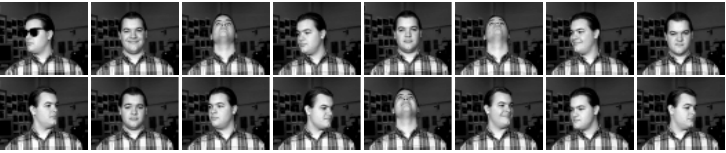}
\vspace{-2ex} 
\caption{An example of images from a person in the CMUFace graph data, where the first image is labeled as an outlier due to the sunglasses.} \vspace{-4ex} \label{fig:CMU1Per}
\end{figure}

Since most network datasets (presented next) lack ground-truth subnetworks, we conduct an experiment on the CMUFace image data (http://archive.ics.uci.edu) since it allows us to evaluate the relevance of uncovered subnetworks \textit{via visualization}. Though images do not originally involve explicit network structures, studying them as graphs has been extensively studied and deemed advantageous~\cite{Shu13}. In particular, it enables the discovery of local image properties, especially in the studies of image denoising and image forensics where pixels can be missing or purposely tampered. Following~\cite{Shu13}, we first down-sample the number of pixels to $50\%$ and construct a common network topology relying on the remaining pixels. Within each image, a pixel corresponds to a node and has edges connected to the 5 nearest pixels. The value associated with a node is the grey level of the corresponding pixel. In order to evaluate whether any method can deal with the heterogeneity in the network dataset, we select all networks with open-eye images from each person as inliers, and randomly select one with sun-glass from any of 4 poses (straight/up/left/right) as an outlier (images from a random person is depicted in Fig.\ref{fig:CMU1Per}). This results in $303$ regular network samples and $20$ anomalous ones, each containing $1,920$ nodes and $11,172$ edges. Subnetworks extracted from sun-glass areas are therefore the ground truths.

\begin{figure}[t]
\hspace{-0.5cm}
\centering
\includegraphics[width=0.9\linewidth]{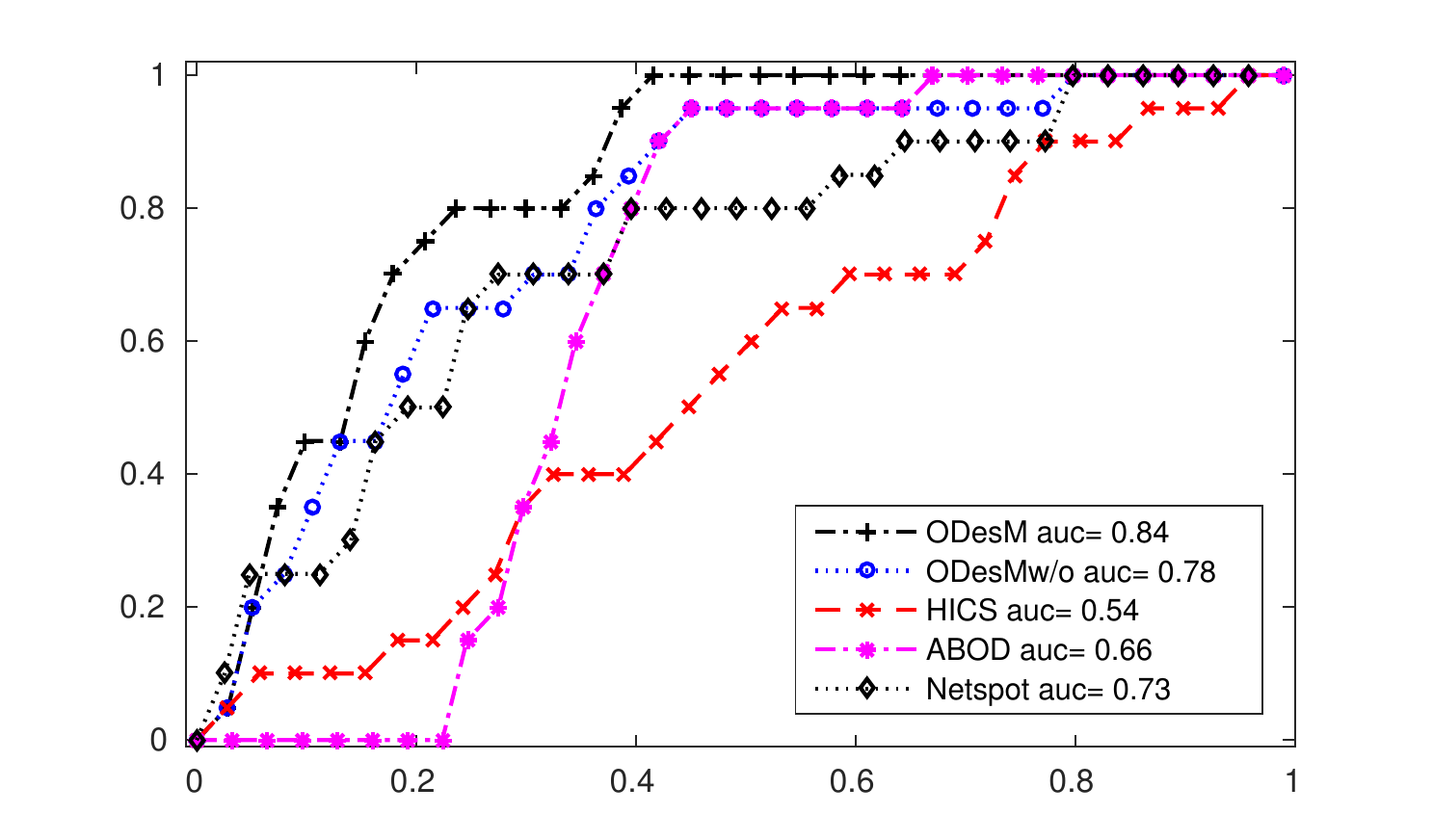}
\vspace{-2ex} 
\caption{ROC curve performance of all algorithms on identifying outlier networks from the CMUFace graph dataset.} \vspace{-4ex} \label{fig:cmu_AUC}
\end{figure}

\textit{Outlier identification:} In Fig.\ref{fig:cmu_AUC}, we plot the ROC curve performance of all algorithms. As seen from this figure, both techniques HiCS and ABOD designed for high dimensional data perform moderately well on this dataset. ABOD explores the variance over angles between an outlier candidate and every pair of other two samples, so its approach explores global outliers deviated from a single distribution of inliers. For this dataset, however, we have multiple distributions. These local outliers are thus harder to be explored by solely relying on the variances of high dimensional vectors' angles. This might explain for the low success rate of ABOD. HiCS, on the other hand, while being designed to find outliers based on contrast subspaces, also does not perform well in this dataset. HiCS attempts to find most information subspace from bottom-up approach and it starts with those of 2-dimension (from a pool of ${1920 \choose 2}=1844160$ possible subspaces).  
If such low dimensional subspaces are not well sampled, it becomes much harder to ensure the most contrast subspaces will be found in higher dimensional subspaces. This is because HiCS retains only 100 to 1000 subspaces in order to avoid the exponential complexity. Netspot performs better than these two techniques by relying on the p-value defined at each node in order to explore significant anomalous regions. However, by converting to a p-value, Netspot also removes the contrast among node's values and thus is less successful in seeking the most potential seed-nodes. Over all techniques, ODesM's performance yields the best with its AUC achieving 0.84, as compared to 0.78 obtained by the second best ODesMw/o. This large gap in AUC clearly confirms the key role of network topology exploited by ODesM, which not only helps it to narrow down the search space of all subgraphs, but also converges to the most explanatory subnetwork structures.

\begin{figure}[t]
\centering
\includegraphics[width=.9\linewidth]{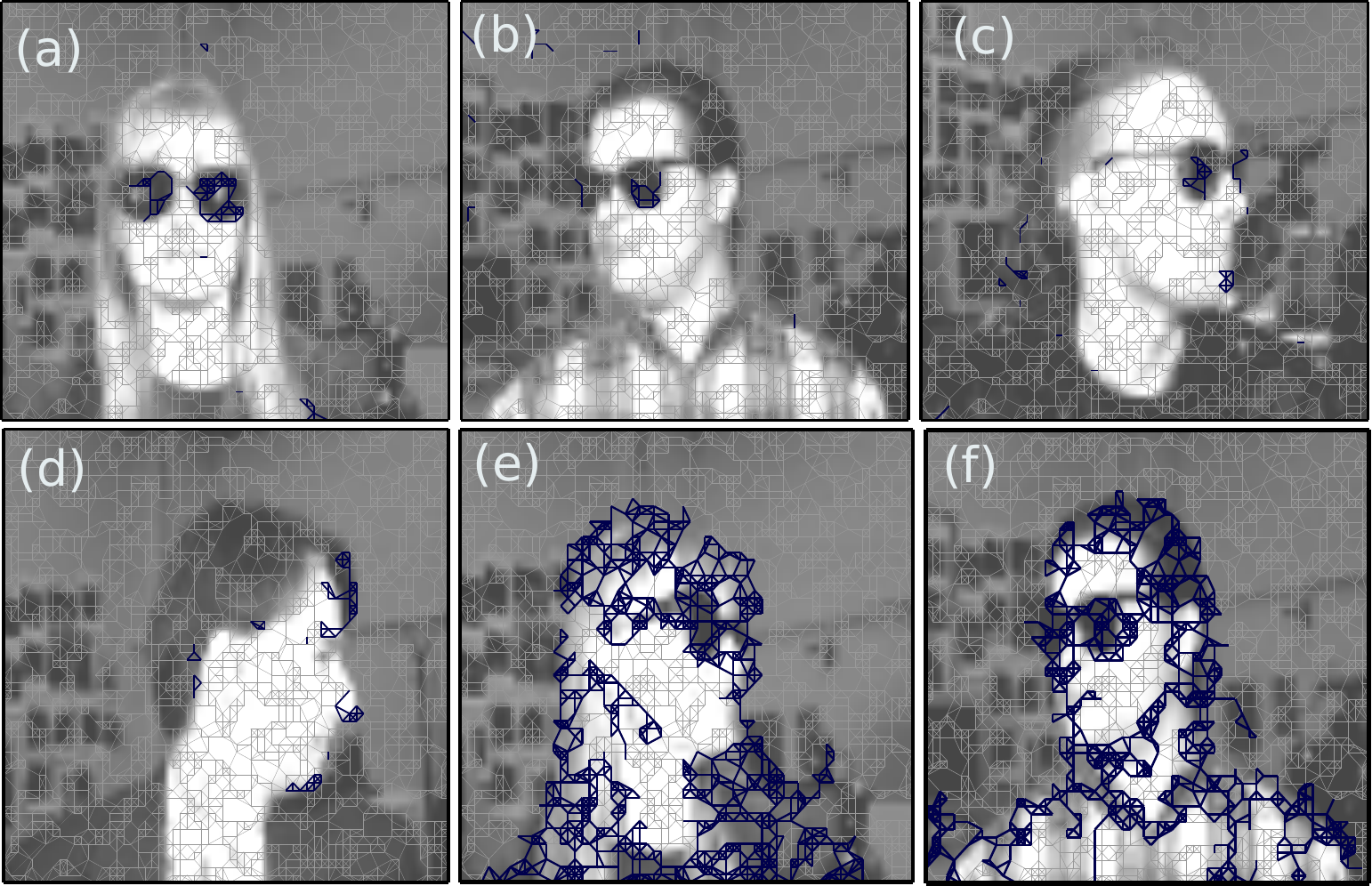}
\vspace{-2ex} 
\caption{Subnetworks selected by ODesM ((a)-(d)) and Netspot ((e)-(f)) in the CMUFace network dataset (detailed explanation is given in text). In each figure, the network topology is shown in grey and the selected subnetworks are shown in blue, while the corresponding full image is shown in background with dimmed colors to improve the visualization (figures are best seen in colors)} \vspace{-4ex} \label{fig:CMUsubnet}
\end{figure}

\textit{Explanatory subnetworks:} We further explore the set of subnetworks discovered by ODesM as the explanation for top ranking outliers. Out of top 20 anomalous networks, 8 are true outliers. We plot in Fig.\ref{fig:CMUsubnet}(a-c) the three top ranked networks that are also truly labeled as outliers and their corresponding images from different poses. In each picture, the full image is shown in background (with dimmed color to boost visualization). The entire network topology is plotted in grey while we color the explanatory subnetworks discovered in blue. As observed, despite coming from different poses, the outlier networks are still well-identified and the subnetworks located around the sunglasses are appropriately selected by ODesM. By visualization, these discriminative substructures clearly explain why an anomalous network is exceptional from regular ones, though they can vary across different outlier networks. We plot in Fig.\ref{fig:CMUsubnet}(d) a network sample that is also ranked high by ODesM but not a true outlier according to the sunglasses' labeling. However, by inspecting its discovered substructures, they still reflect some exceptional property of this image, where all subnetworks have been selected at the curve of the face. Generally, such kind of substructures are quite typical for each individual person.

\Comment{
\footnote{It is worth mentioning that these regions and other ones can be clustered by an image partitioning algorithm. However, the most relevant and discriminative ones can be found only by contrasting them against other samples, which is usually not addressed in image partitioning. 
}
}

Recall that ABOD, ODesMw/o and HiCS are not network-based techniques. While ABOD identifies outliers based on variance of vector angles, ODesMw/o selects individual nodes and does not explore subneworks. HiCS generates multiple subspaces for a single outlier candidate and there is no obvious way to derive subnetworks from all of them. Hence, we select Netspot for comparison based on its discovered anomalous subnetwork regions. In Fig.\ref{fig:CMUsubnet}(e-f), we plot two typical true outliers found from 20 top networks ranked by Netspot based on the anomalous score of the selected subnetwork regions. 
It can be seen that, unlike the subnetworks discovered by our method, it is hard to justify why the corresponding images are exceptional though they are strongly connected. 

In both figures, the substructures from entire faces have been selected. This performance probably comes from the fact that, other than p-value, Netspot also relies on the adjacency of network samples to derive the time interval at which significant anomalous regions can appear. However, once the interval is set to 1 (i.e. for each individual network), it has limited information to justify the relevance of a network region since there is no temporal development among network samples. Thus, the p-value computed at each node is likely playing the key role. And as long as its values do not change abruptly, Netspot tends to select all of them, forming a large subnetwork structures as shown in Fig.\ref{fig:CMUsubnet}(e-f). The patterns discovered between Netspot and our ODesM are thus fundamentally different. For this reason, we do not attempt to compare their uncovered subnetworks in the subsequent experiments.

\subsection{Biological PPI network}                  \label{sec:CMU}

\begin{figure}[t]
\hspace{-0.5cm}
\centering
\includegraphics[width=0.9\linewidth]{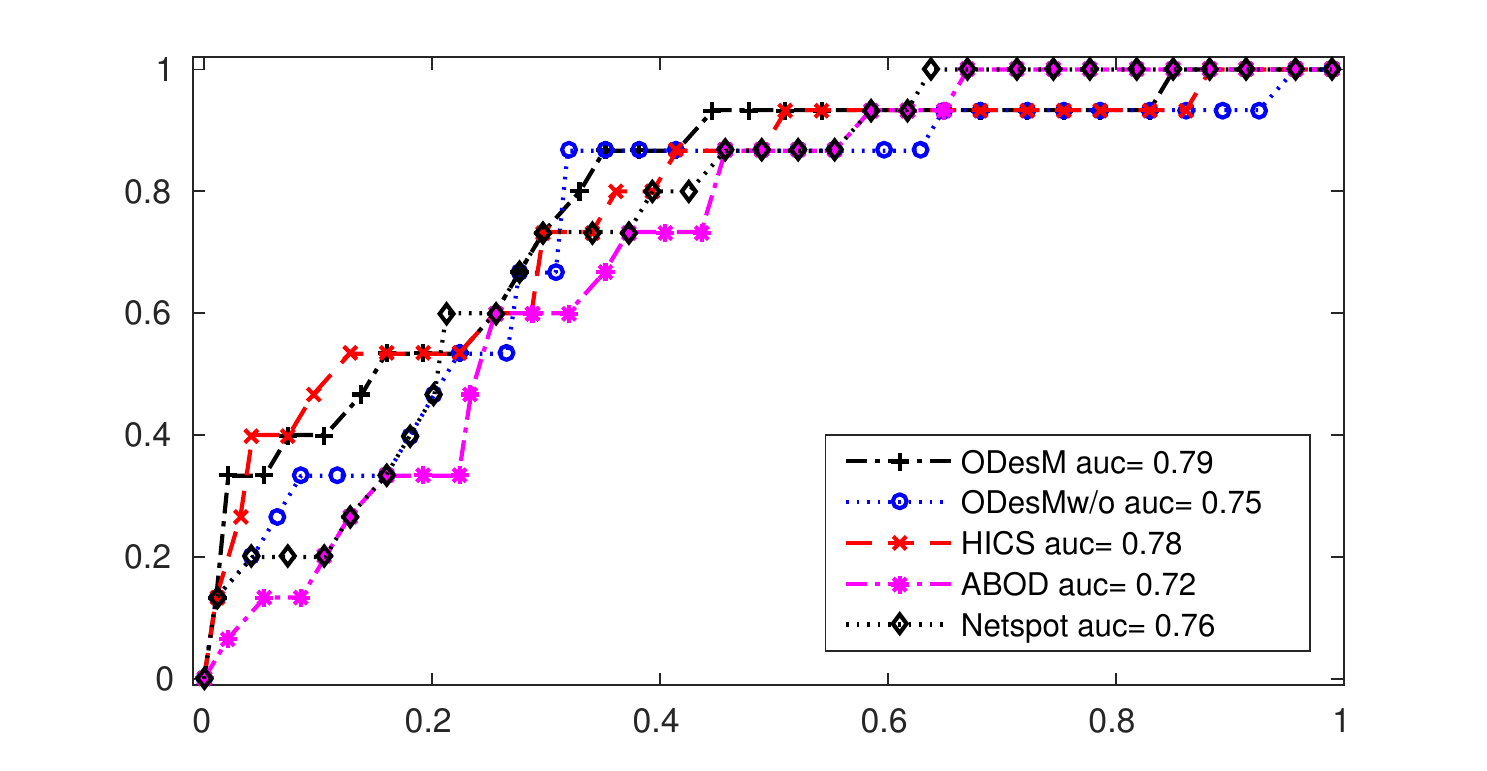}
\vspace{-2ex} 
\caption{ROC curve performance of all algorithms on identifying outlier networks from the Liver gene network dataset.} \vspace{-2ex} \label{fig:Liver_AUC}
\end{figure}

The second dataset we use for evaluation is the Liver metastasis in human~\cite{Dong07} with the gene network derived from the protein-protein interaction. Values associated with nodes are the gene expression values. The dataset contains $7,383$ genes and $251,916$ edges collected from 101 healthy subjects viewed as inlying network samples, and 15 diseased subjects labeled as outliers.

\textit{Outlier identification:} We show in Fig.\ref{fig:Liver_AUC} the ROC curve of all algorithms on the Liver dataset. The performance of our ODesM method is competitive to that of HiCS and both are better than the remaining techniques. Netspot also performs well on this dataset as indicated by its 0.76 AUC value and slightly better than ODesMw/o.
Recall that each network sample of this dataset also contains a large number of nodes. However, unlike the CMUFace graph data where we have multiple data distributions (each representing images from an individual person), here we have only a single network distribution of healthy subjects. The outlier prediction rates of all techniques are thus not as diverse as those we have seen in the CMUFace graph dataset. However, the results still indicate that our ODesM algorithm yields the highest outlier prediction rate. 

\begin{figure}[t]
\centering
\includegraphics[width=.9\linewidth]{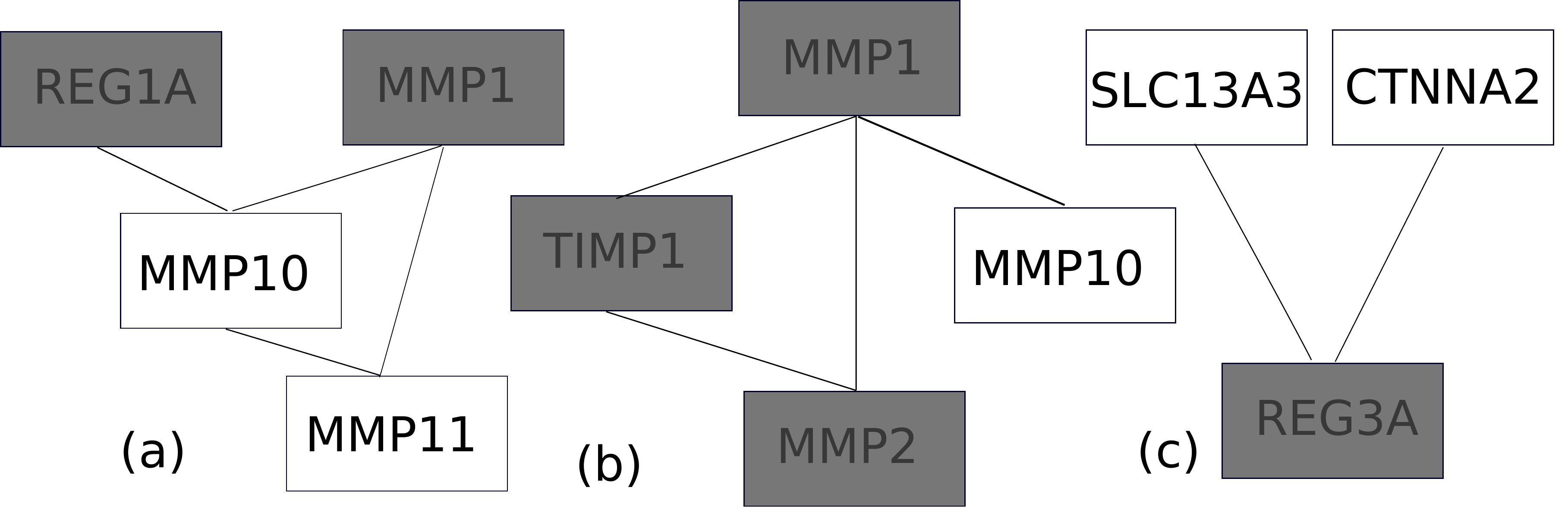}
\vspace{-2ex} 
\caption{Subnetworks frequently discovered by ODesM in its top 15 network samples with the highest outlier scores. Shaded genes are related to the liver metastasis cancer.} \vspace{-4ex} \label{fig:liverSubgraph}
\end{figure}

\textit{Explanatory subnetworks:} There are no obvious ground truths for the gene pathways (subetworks) associated with the liver cancer. However, as an attempt to investigate how relevant and explanatory are the subnetworks discovered by ODesM, we compute the most frequent subnetworks found in the top 15 ranked outlying networks. In Fig.\ref{fig:liverSubgraph}, we plot 3 subgraphs that have the highest frequency. The first subnetwork is found in 6 networks and out of these, 4 are anomalous networks. The second subnetwork is found in 4 networks with 3 as true outliers. Among these, two discovered subnetworks, the genes REG1A, MMP1, MMP2 and TIMP1 (shaded in the Fig.\ref{fig:liverSubgraph}) are particularly interesting since they are in agreement with the ones found in~\cite{Dong07} and have been reported to be involved in liver metastasis. The last subnetwork is found in 5 network samples and among them, only one is a true outlier. Though the genes forming the above subnetworks are not all related to the liver cancer and not all diseased subjects are ranked at the top (7 true outliers are found out of top 15), an important observation from these results is that, the frequent involvement of diseased genes in the discovered subnetworks can signal the appearance of the disease. Moreover, since diseased subjects can suffer from different stages or subtypes of the cancer, the disease-related gene pathways can possibly vary from one subject to another. These uncovered subnetworks thus do carry explanatory information to justify why an unhealthy subject is an outlier.

\subsection{Road traffic networks}                  \label{sec:Traffic}

The last dataset we use for evaluation is LATraffic---the highway traffic network data of Los Angeles, California (http://pems.dot.ca.gov) during April 2011. LATraffic contains multiple network snapshots of size of 100/128 nodes/edges. Each node in the network corresponds to a road segment and its associated value is the average vehicle speed within 5-minute resolution. In generating outlier labels for the network samples, we rely on the distribution of the average speed computed for each snapshot. Specifically, 300 snapshots are randomly selected around the mean of this distribution and labeled as regular networks. Other 30 snapshots are randomly selected from two extreme tails (15 each) of this distribution and labeled as anomalous networks.

\begin{figure}[t]
\hspace{-0.5cm}
\centering
\includegraphics[width=0.9\linewidth]{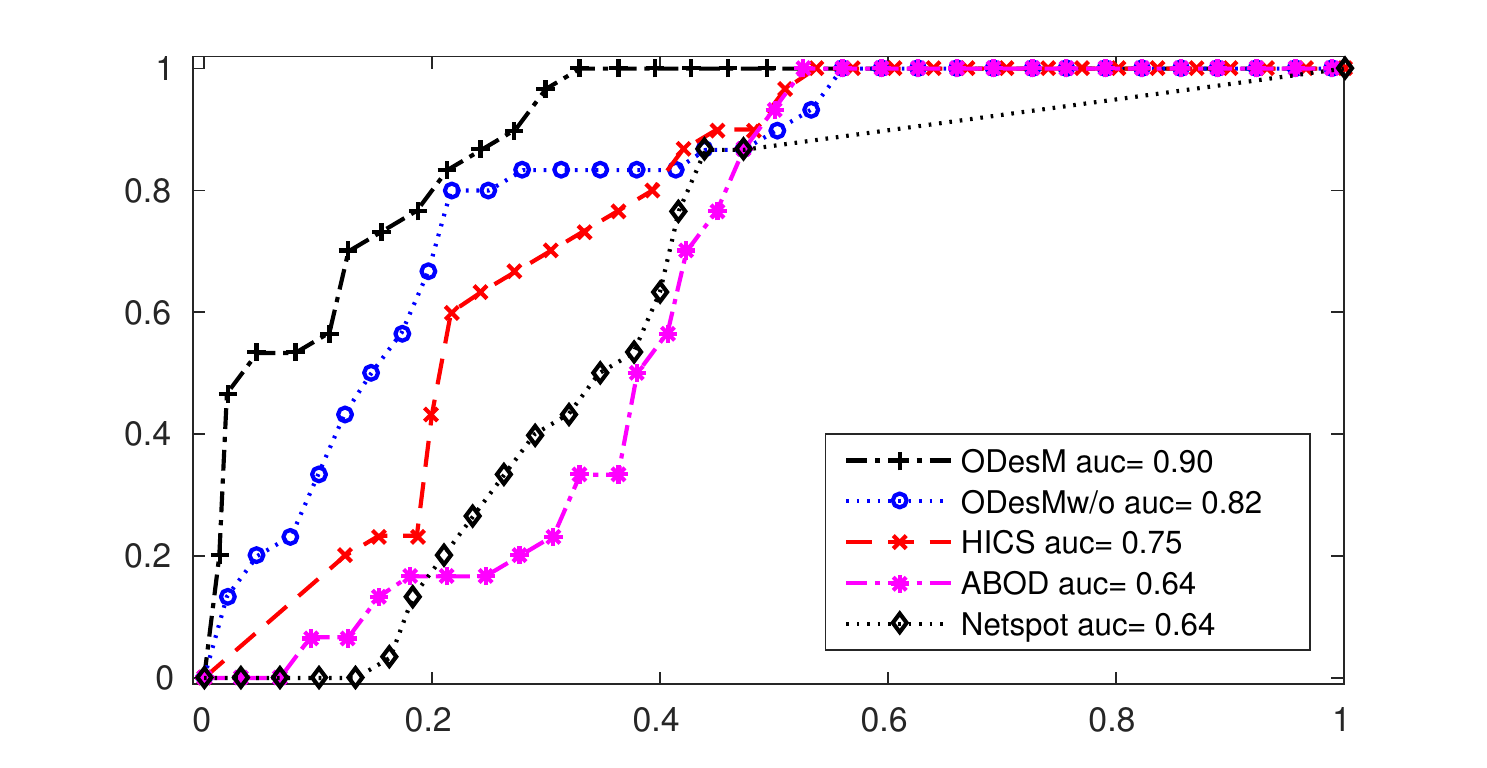}
\vspace{-2ex} 
\caption{ROC curve performance of all algorithms on identifying outlier networks from the LATraffic network dataset.} \vspace{-4ex} \label{fig:TraffS_AUC}
\end{figure}

\textit{Outlier identification:} The ROC curve performance of all algorithms on the LATraffic is shown in Fig.\ref{fig:TraffS_AUC}. For this relatively small network, HiCS handles the subspace candidates well and its Monte-Carlo sampling based approach tends to select high contrast subspaces. Regarding the performance of Netspot, recall that 
the dataset contains two types of outliers, one with high average speed and the other with low speed. By relying on the notion of network fraction in computing the p-value for each node, Netspot may not be able to find both types of outliers. Among all examined techniques, ODesM is still the best performer with its AUC score at 0.9. 
Deeper investigation on its outlier ranking further shows that ODesM predicts 16 out of top 20 networks as true outliers and they are from both low and high average speeds.

\begin{figure}[t]
\centering
\includegraphics[width=1\linewidth]{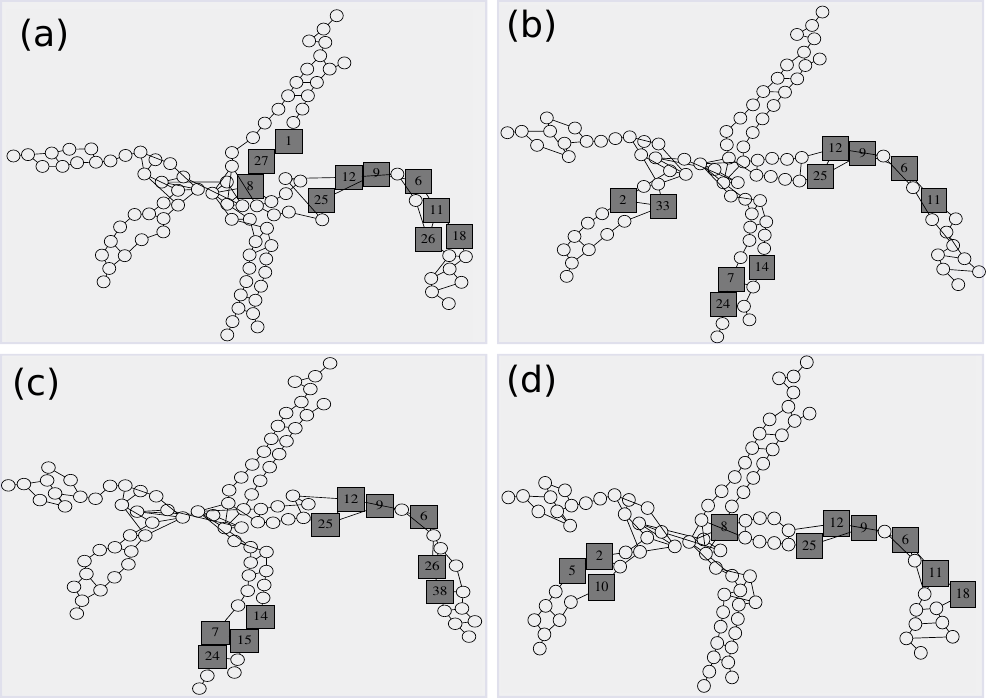}
\vspace{-2ex} 
\caption{Top 4 outlier networks discovered by ODesM from the LATraffic. Two networks shown in (a) and (d) are from the low speed distribution while the two shown in (b) and (c) are from the high speed distribution. Road segments involved in the explanatory subnetworks are shaded.} \vspace{-5ex} \label{fig:TraffSubnet}
\end{figure}

\textit{Explanatory subnetworks:} We further explore the set of subnetworks discovered by ODesM for its top ranking network snapshots. In Fig.\ref{fig:TraffSubnet}, we plot the uncovered subnetworks for top four outlier networks. The networks in (a) and (d) are the true outliers with low speed while the ones in (b) and (c) are the true outliers with high speed. The sets of discovered subnetworks in both cases are quite consistent. Taking a closer look of these explanatory substructures, there is an interesting point to highlight. Apparently, we would expect the explanatory subnetworks for two types of outliers to be different since one was chosen from the low speed distribution while the other one was selected from the high speed distribution. However, it turns out that they share one large subnetwork spanned by the nodes 11,6,9,12 and 25. The common selection of this subnetwork in both kinds of outliers may suggest that such a set of adjacent road segments is highly sensitive to the traffic congestion. For monitoring purposes, these road segments should be the top candidate to be selected since they are likely to reflect the overall condition of the entire traffic network.

\subsection{Impact of parameters}\label{ParamScalability}

\begin{figure}[t]
\centering
\includegraphics[width=.7\linewidth]{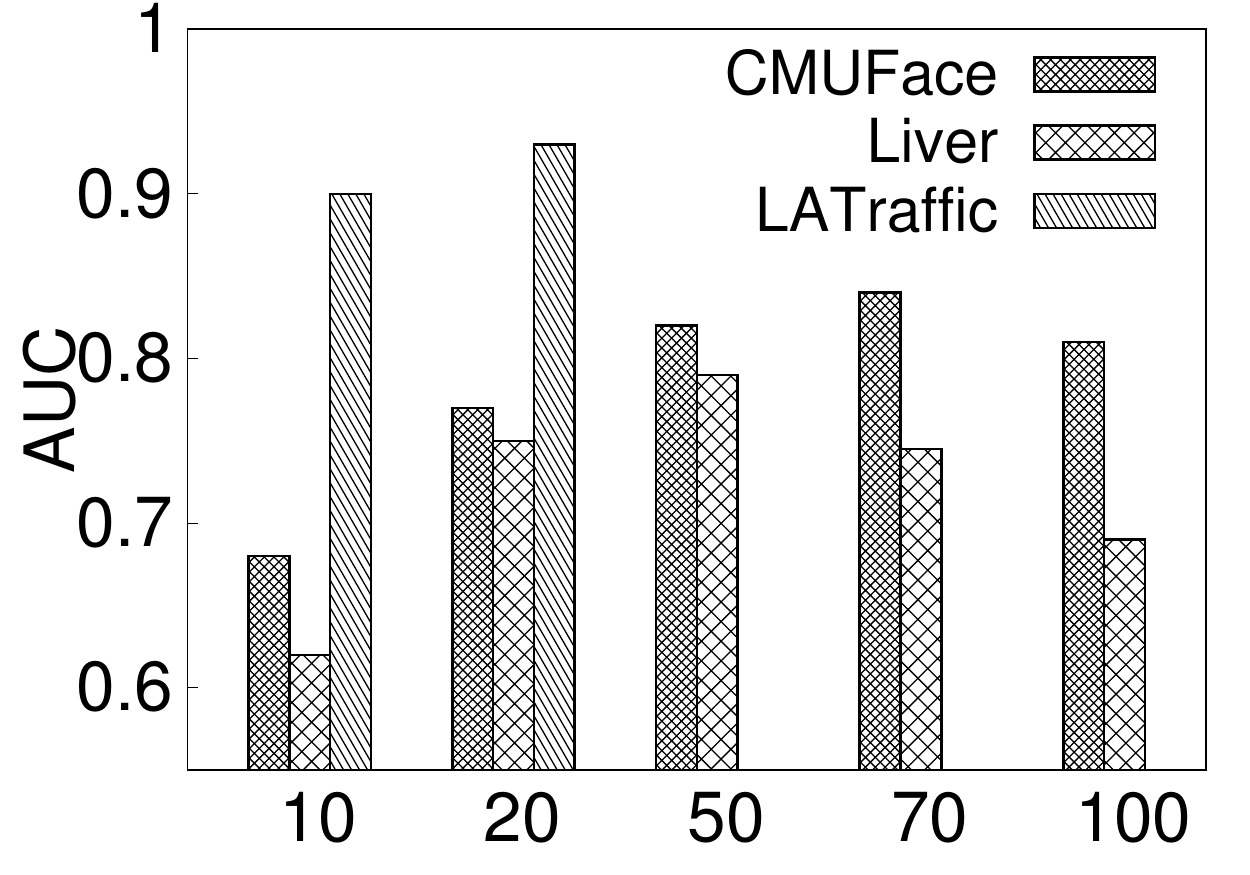}
\vspace{-2ex} 
\caption{The AUC Performance of ODesM by varying the subnetwork sizes between 10 to 100 nodes. For LATraffic, the subnetwork size is limited to 20 nodes since the entire network is small with only 100 total nodes..} \vspace{1ex} \label{fig:AUCvsSelFea}
\vspace{-4ex}%
\end{figure}

ODesM requires three parameters to be set: $K$ determining the number of network neighbors, $\lambda_1$ deciding the influence of network topology, while $\lambda_2$ controlling the discovered subnewtork size. As discussed in Section \ref{Sec:Discussion}, 
we select a range of values for $K$ and $\lambda_1$ and apply the best-effort-approach~\cite{LOF,Arthur12} to compute outlierness for each network sample. In this experiment, we thus only report the impact of varying $\lambda_2$ on the performance of outlier detection. Since our three datasets are vastly different in network size, we will use specific values to limit the size of selected subnetworks. In Fig.\ref{fig:AUCvsSelFea}, we report the AUC performance by varying the total number of nodes for the discovered subnetworks between 10 to 100. For the LATraffic network data, we do not consider the subnetwork size larger than 20 since the whole network has only 100 nodes.

\Comment{
\begin{table}[t]
\begin{center}
\begin{tabular}{rrrrrr}
\hline
   DS &    10 &     20 &   50 &    70 &      100 \\
\hline
    CMUFace &  0.68&      0.77 &       0.82 &       0.84&       0.81 \\

    LATraffic &  0.9 &      0.93 &       0.0 &       0.0&       0.0 \\

    Liver &     0.62 &     0.75 &       0.79 &       0.75&       0.69\\
\hline
\end{tabular}
\caption{AUC Performance of ODesM by varying the subnetwork sizes between 10 to 100 nodes. For LATraffic, the subnetwork size is limited to 20 nodes since the entire network is formed from 100 nodes.} \label{tab:AUCvsSelFea}
\end{center}
\end{table}
}

A general trend can be observed from Fig.\ref{fig:AUCvsSelFea}. As the total number of nodes for subnetworks becomes larger, the outlier detection rate tends to increase. However, for Liver and CMUFace datasets, when the discovered subnetworks are larger than 70 nodes, the outlier detection rates get reduced. This might happen since choosing larger values for the subnetworks may further include irrelevant substructures, which leads to a higher rate of false positive prediction.

\section{Related work}                    \label{sec:Related-work}

Outlier detection from network data can generally be divided into two categories: those addressing plain networks~\cite{Ding12,Oddball10} and those focusing on attributed networks~\cite{Gao2010,Perozzi14}. In the first category, only information about the network topology is available and most studies adopt structure-based~\cite{Ding12,Oddball10} and community-based methods~\cite{Sun05} to spot nodes or small groups of nodes that have abnormal connectivity patterns. In the second category, attributes associated with nodes and edges are also available. Discovering outlying patterns therefore seeks not only abnormal connectivity structure but also coherence of network attributes~\cite{Gao2010,Perozzi14}. Network properties like normality~\cite{Leman16}, conductance~\cite{conductance06} and Oddball~\cite{Oddball10}
are often employed to quantify the internal consistency and external separability (collectively anomalous degree) of a set of nodes (local communities). Most of these studies focus on searching outlying patterns from a single network, which contrasts with our work that addresses the problem in a more general setting of multiple networks. Several recent studies~\cite{Chen12,Aggarwal12,GuptaGSH12,Netspot13} developed for dynamic networks are closer to ours. In~\cite{Chen12}, the authors present 6 types of community-based outliers including shrink, grow, merge, split, born and vanish. Such types of anomalous communities can be identified via tracking the evolution of communities over time. In~\cite{Aggarwal12}, the temporal distribution of the number of messages exchanged in a social network (like Twitter) is used as means to detect abnormal events. More specifically, if the fraction of edges added to a community with the current time window is significantly larger than the previous one, then it can signal that a special event is within that community. Authors in~\cite{MEDEN11} introduce a novel problem of mining a heaviest dynamic subgraph (HDS) in a time evolving network. The problem is shown NP-hard, and a heuristic algorithm named MEDEN is developed based on the filter-and-verify framework. This study is recently extended in~\cite{Netspot13} to the NetSpot technique that enables the mining of multiple HDSs. NetSpot approximates HDSs via a local search approach and it alleviates the local optimal solutions via exploring a large range of neighborhood search\cite{Netspot13}.
Other studies \cite{Aggarwal11} monitor global network parameters/probabilities to detect events/changes while those developed in~\cite{GuptaGSH12} attempt to spot anomalous nodes and edges. They are thus less relevant to our studies. In contrast, we do not focus on searching for outlying patterns from a single dynamic evolving network but from multiple network samples. Moreover, our focus is on discovering outliers as entire network samples but localizing subnetworks to explain why such network samples are exceptional.

Outlier detection in high dimensional spaces~\cite{Arthur12} can also be conceptually related to our studies. Two popular approaches to deal with this problem are from subspace sampling~\cite{HiCS} and subspace projection~\cite{COP}. Techniques from subspace sampling generally assume that outliers only show up in low dimensional subspaces and such subspaces can be discovered via sampling combined with relevant statistical tests. In contrast, methods based on space transformation directly search for a single subspace, often a linear combination of all original features, that maintains certain properties, e.g. variance, of the data. Outliers can be found from this induced low dimensional subspace. 
Though these techniques are effective in ranking and finding anomalous objects, directly applying them to network data often lacks domain relevance since the nature of mutual interaction among network entities is completely ignored. Additionally, while a novel subspace is effective in computing outlier scores, it barely provides qualitative explanation for each individual outlier.

\section{Conclusions} \label{sec:sew:con}

In this paper, we addressed an important problem of identifying and explaining outlier network samples. A novel algorithm was developed to identify subnetworks that discriminate outlier networks from their neighboring regular network samples. The algorithm was designed in the framework of network regression combined with the constraint on the network topology and the L1-norm shrinkage to perform subnetwork discovery. Our algorithm thus goes beyond both subspace learning and subgraph discovery methods by directly learning the most discriminative subnetworks to justify the exceptional properties of an anomalous network. Evaluation on various real-world network datasets demonstrated that our novel algorithm not only outperformed existing techniques, but also uncovered highly relevant and interpretable local subnetworks.

As future work, we would like to extend our research to handle databases with very large networks. Obviously, directly applying ODesM might not be highly scalable as analyzed in Section \ref{Sec:Discussion}. To deal with very large networks, we could apply network compression~\cite{graphCompression08} that allows us to summarize both network topology and signals on the nodes. This is equivalent to representing a large network at different scales/resolutions. The open research issues are therefore: (i) How can we trade-off between the size of compressed networks, in exchange for scalability, and the quality of outlier detection? (ii) How can we ensure that the most exceptional information (explaining for an outlier network) is not compromised by such a compression approach?

\bibliographystyle{abbrv}
\bibliography{ArXivODeSM}

\end{document}